\documentclass[conference]{IEEEtran}
\IEEEoverridecommandlockouts
\usepackage{cite}
\usepackage{amsmath,amssymb,amsfonts}
\usepackage{algorithmic}
\usepackage{multirow}
\usepackage{graphicx}
\usepackage{hyperref}
\usepackage{subfigure}
\usepackage{textcomp}
\usepackage{xcolor}
\usepackage{enumerate}
\usepackage{comment}
\makeatletter
\newcommand{\linebreakand}{%
  \end{@IEEEauthorhalign}
  \hfill\mbox{}\par
  \mbox{}\hfill\begin{@IEEEauthorhalign}
}
\makeatother
\def\BibTeX{{\rm B\kern-.05em{\sc i\kern-.025em b}\kern-.08em
    T\kern-.1667em\lower.7ex\hbox{E}\kern-.125emX}}
\begin{document}

\title{Automating lichen monitoring in ecological studies using instance segmentation of time-lapse images}

\author{
\IEEEauthorblockN{\hspace{-1cm} Safwen Naimi, Olfa Koubaa, Wassim Bouachir}
\IEEEauthorblockA{\textit{\hspace{-1cm}Data Science Laboratory}\\ 
\textit{\hspace{-1cm}University of Quebec (TÉLUQ)}\\
\hspace{-1cm}Montréal, QC, Canada}
%email address or ORCID
\and
\IEEEauthorblockN{Guillaume-Alexandre Bilodeau}
\IEEEauthorblockA{\textit{LITIV lab.} \\
\textit{Polytechnique Montréal}\\
Montréal, QC, Canada}
%email address or ORCID
\linebreakand
\IEEEauthorblockN{\hspace{1.6cm}Gregory Jeddore}
\IEEEauthorblockA{\textit{\hspace{1.6cm}Miawpukek First Nation} \\
\textit{\hspace{1.6cm}Natural Resources Canada (NRCan)}\\
\hspace{1.6cm}Conne River, NL, Canada}
%email address or ORCID
\and
\IEEEauthorblockN{\hspace{1.1cm}Patricia Baines, David L. P. Correia, André Arsenault}
\IEEEauthorblockA{\textit{\hspace{1.1cm}Canadian Forest Service}\\
\textit{\hspace{1.1cm}Natural Resources Canada (NRCan)}\\
\hspace{1.1cm}Corner Brook, NL, Canada}
%email address or ORCID
}

\maketitle

\begin{abstract}
Lichens are symbiotic organisms composed of fungi, algae, and/or cyanobacteria that thrive in a variety of environments. They play important roles in carbon and nitrogen cycling, and contribute directly and indirectly to biodiversity. Ecologists typically monitor lichens by using them as indicators to assess air quality and habitat conditions. In particular, epiphytic lichens, which live on trees, are key markers of air quality and environmental health. A new method of monitoring epiphytic lichens involves using time-lapse cameras to gather images of lichen populations. These cameras are used by ecologists in Newfoundland and Labrador to subsequently analyze and manually segment the images to determine lichen thalli condition and change. These methods are time-consuming and susceptible to observer bias. In this work, we aim to automate the monitoring of lichens over extended periods and to estimate their biomass and condition to facilitate the task of ecologists. To accomplish this, our proposed framework uses semantic segmentation with an effective training approach to automate monitoring and biomass estimation of epiphytic
lichens on time-lapse images. We show that our method has the potential to significantly improve the accuracy and efficiency of lichen population monitoring, making it a valuable tool for forest ecologists and environmental scientists to evaluate the impact of climate change on Canada's forests. To the best of our knowledge, this is the first time that such an approach has been used to assist ecologists in monitoring and analyzing epiphytic lichens.
\end{abstract}


\begin{figure*}[t]
\centering
\subfigure[]{\includegraphics[width=0.30\textwidth]{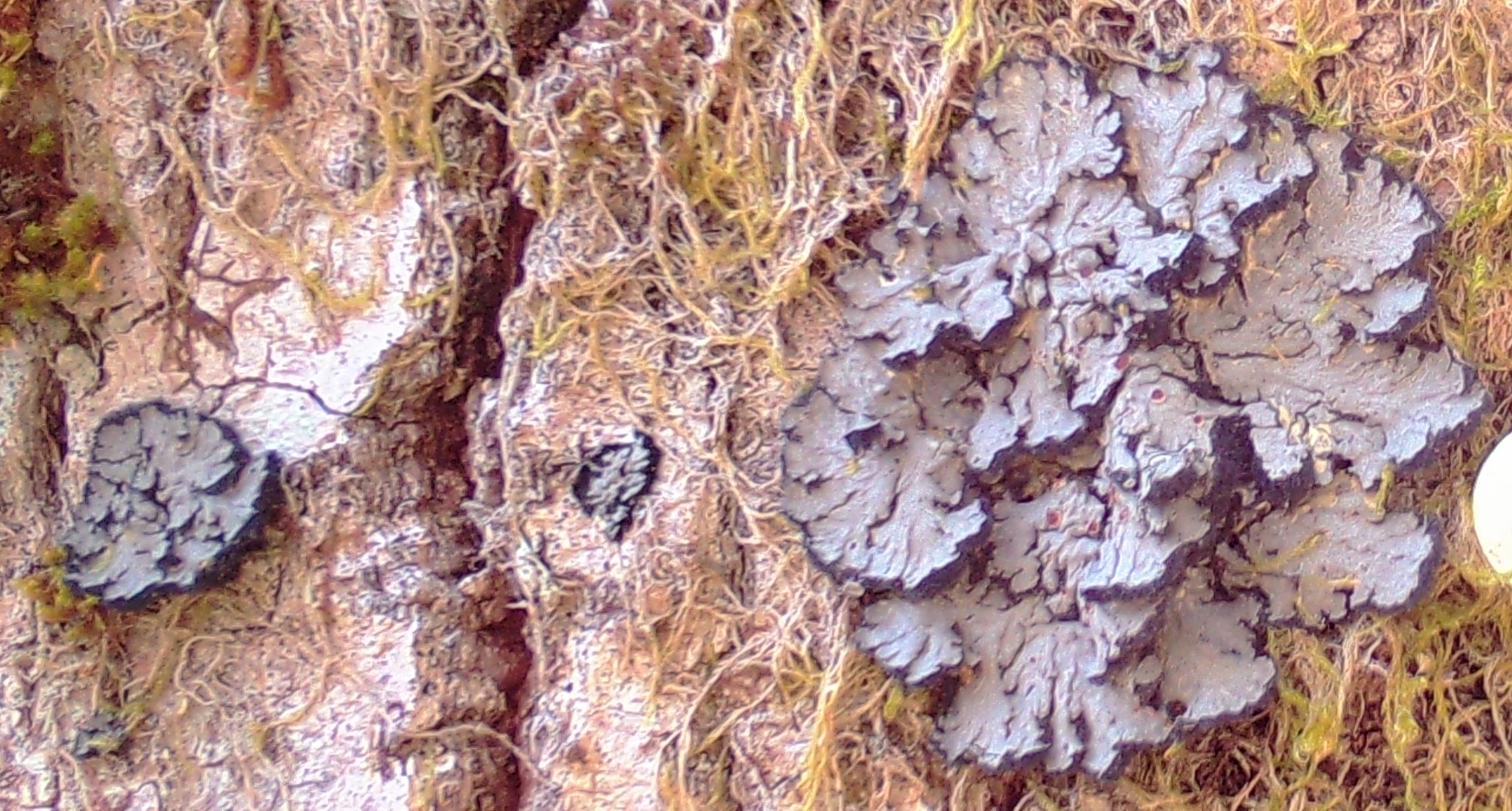}\label{fig:111}}\hspace{5mm}
\subfigure[]{\includegraphics[width=0.29\textwidth]{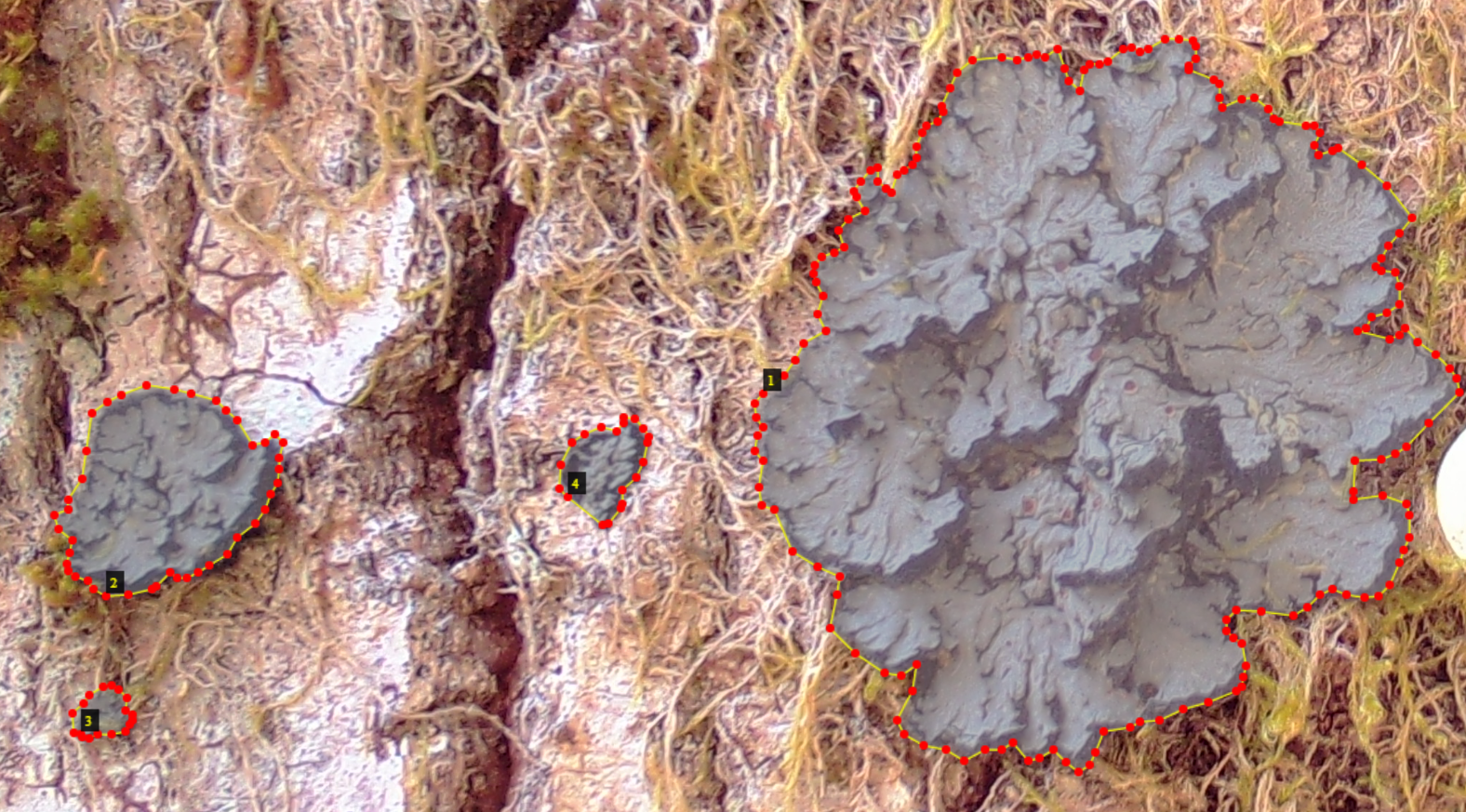}\label{fig:112}}\hspace{5mm}
\subfigure[]{\includegraphics[width=0.28\textwidth]{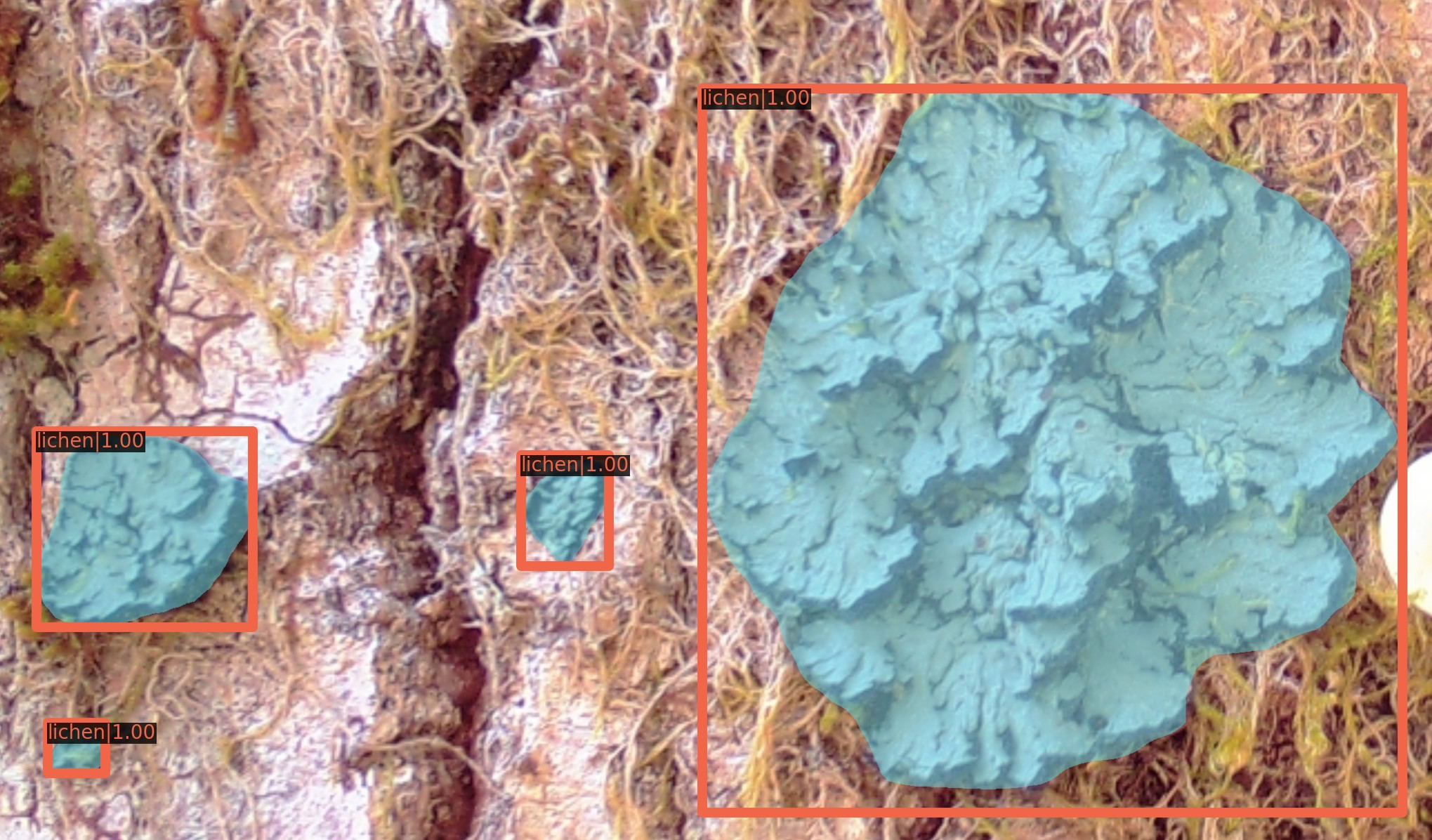}\label{fig:113}}\hspace{5mm}
\caption{The qualitative results of our model (a) Input image (b) Ground-truth (c) Segmentation with our method. We can observe that our method can successfully segment all the lichens with good precision for the boundary shape.}
\label{fig:model}
\end{figure*}

\section{Introduction}
\begin{figure*}[ht]
\centering
\subfigure[]{\includegraphics[width=0.25\textwidth]{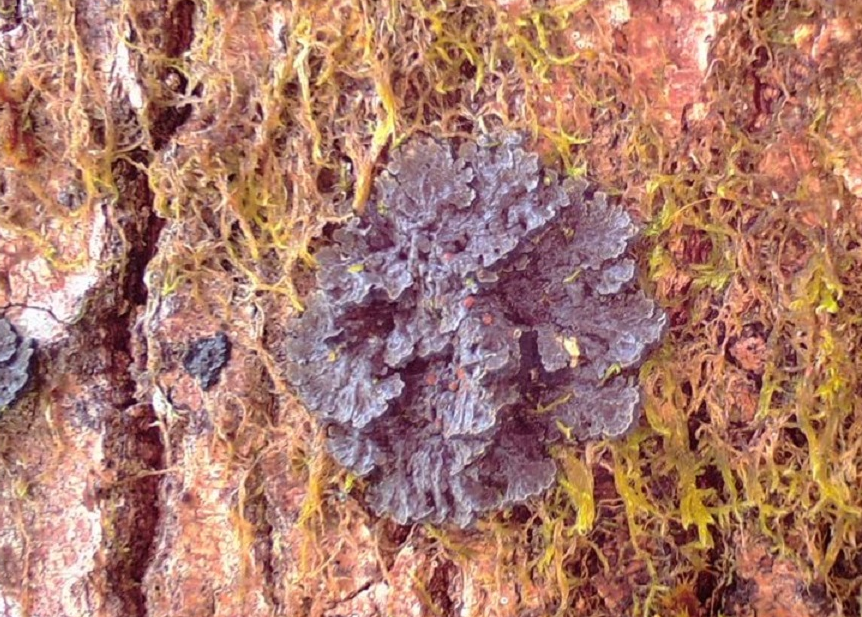}\label{fig:1}}\hspace{5mm}
\subfigure[]{\includegraphics[width=0.25\textwidth]{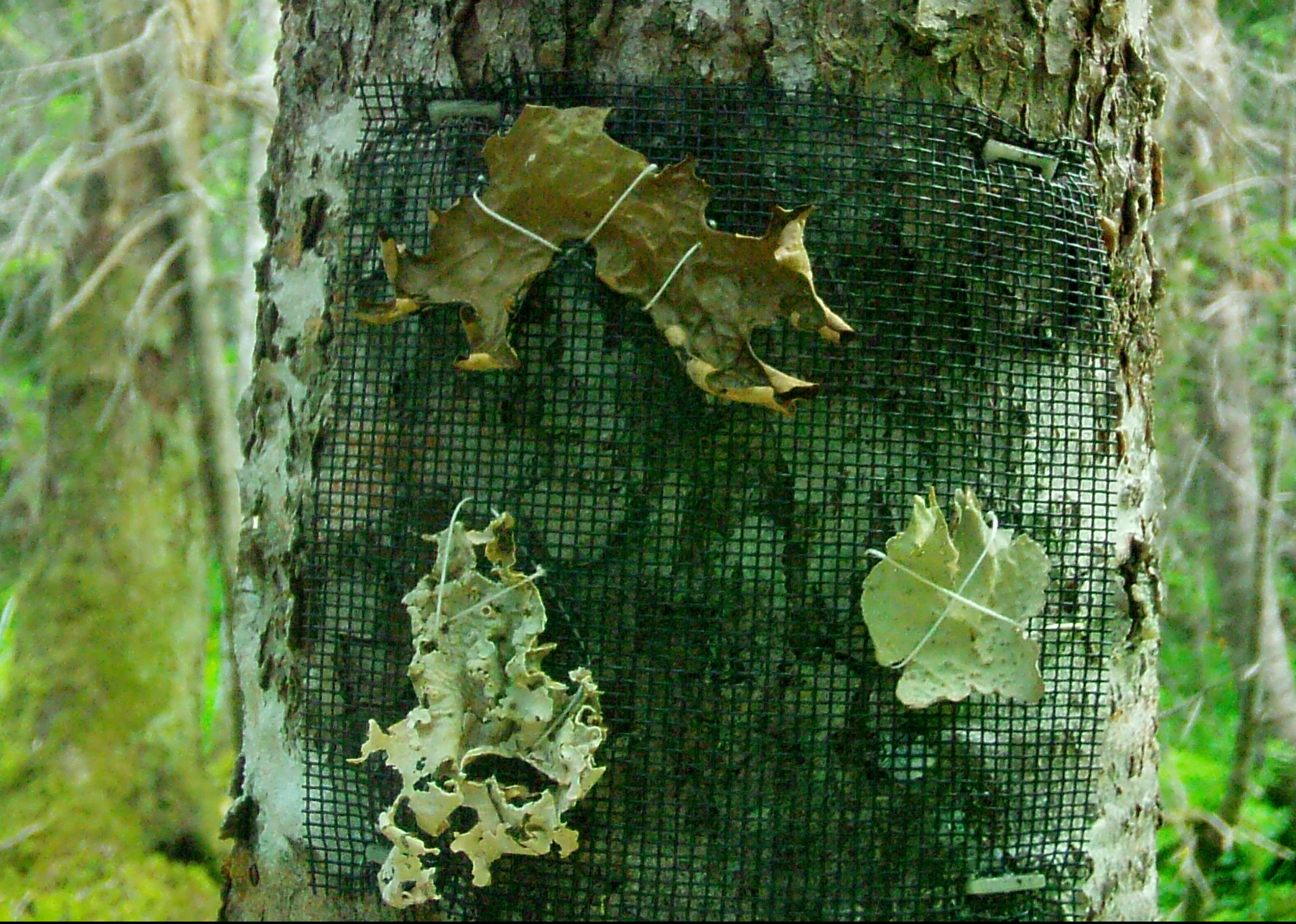}\label{fig:2}}\hspace{5mm}
\subfigure[]{\includegraphics[width=0.25\textwidth]{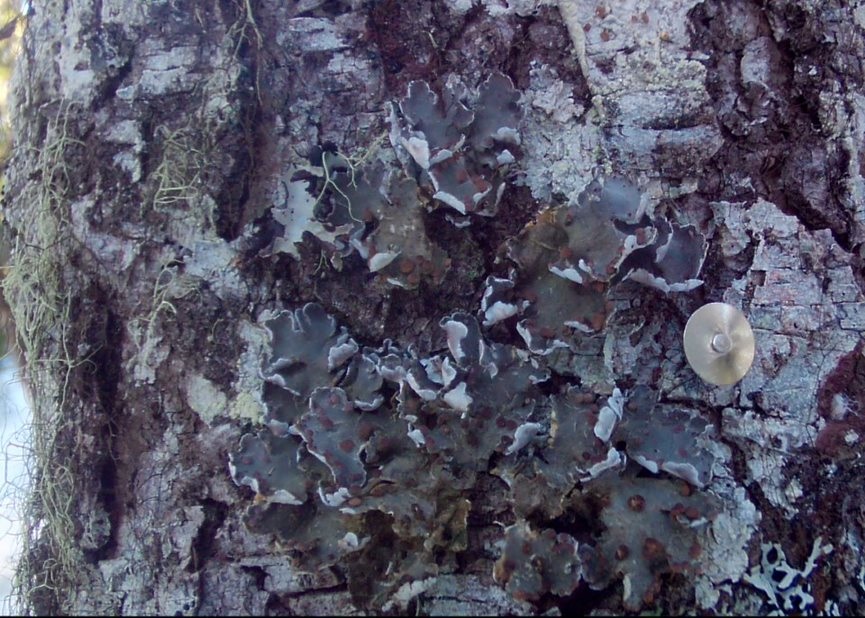}\label{fig:3}}
\caption{Examples of different epiphytic lichens in our dataset. In \ref{fig:1}, we present an image of \textit{Pectenia plumbea}, commonly found in North America and Europe, growing on a variety of substrates including soil, rock, and bark \cite{Otlora2017SpeciesDA}. \ref{fig:2} shows an image of \textit{Lobaria pulmonaria}, a large and leafy lichen that can grow up to 30 cm in diameter \cite{GaioOliveira2004GrowthIR}. In \ref{fig:3}, we present an image of \textit{Erioderma pedicellatum}, which is characterized by its flat, lobed thallus that ranges in color from light green to brownish-gray \cite{Maass1983NewOO, Gauslaa2020TheCE}.}
\label{fig:lichens}
\end{figure*}

Lichens have a remarkable ability to thrive in nutrient-poor environments and colonize previously barren surfaces because of their slow growth rate \cite{Cutler2010LongtermPS}. Lichens also have the capability to persist and flourish in later stages of ecological succession. Furthermore, they play important functional roles in many ecosystems. In fact, lichens are major contributors to the overall carbon and nitrogen cycling, biodiversity, and biomass \cite{Cutler2010LongtermPS}\cite{Elbert2012ContributionOC}. Epiphytic lichens are a type of lichens living on trees, and other plants without harming them. They obtain nutrients from air and rainwater. Epiphytic lichens are important indicators of air quality and environmental health. They also play an essential role in many ecosystems by providing food and shelter to various organisms. They display a range of colors and shapes (see Figure \ref{fig:lichens}), and have practical uses in traditional medicine, dye production, and food.

Ecologists are keenly interested in monitoring epiphytic lichen populations due to various reasons, such as their sensitivity to environmental changes and their role in indicating forest health. Traditional methods for monitoring epiphytic lichen populations are based on field surveys \cite{Aragn2016ASM}\cite{Aragn2013EstimatingEL}, which require physically visiting the site regularly to collect data on lichen abundance and diversity. In these traditional methods, the lichen is identified and the specific measurements are taken such as size, condition, and number of reproductive structures. This approach is often labor-intensive, time-consuming, and may miss subtle changes in lichen populations due to their infrequency. Moreover, the accuracy of the surveys may vary depending on the expertise and attention to detail of the observer. In addition, lichens grow slowly, so changes in their populations may take years to manifest, making it challenging to detect and respond to any emerging issues quickly. They may undergo subtle changes that need near-continuous monitoring to be detected. Climate change and other environmental issues are increasing the demand for efficient and accurate monitoring technologies. 

A new approach for monitoring biodiversity was developed as part of the boreal sentinels project which is co-led by the Canadian Forest Service and Miawpukek First Nation. The purpose of this project is to integrate indigenous knowledge and science to develop a state-of-the-art biodiversity monitoring system with a focus on epiphytic lichens. In particular, in this project, time-lapse cameras mounted on trees were deployed to monitor epiphytic lichens, such as the globally endangered boreal felt lichen in Newfoundland and Labrador in Canada. This camera network allows the recording of a large volume of sequential image data for monitoring lichen evolution over long period of time. Using instance segmentation, we aim to automate the observation of lichens over long periods of time and quantify their biomass. In fact, different species of epiphytic lichens have unique characteristics that can affect their ecological role and interactions with other organisms. Our main contribution is the design of an accurate method that addresses the problem of segmenting epiphytic lichen as shown in Figure \ref{fig:model} to automate monitoring and biomass estimation by ecologists. Our method has also the capacity to recognize new species of lichen not included in the training. To the best of our knowledge, this is the first computer vision method for automating the monitoring of epiphytic lichens.

The rest of this article is organized as follows. Section \ref{related_work} introduces the related works. Section \ref{method} provides a detailed description of the proposed method. Experimental setup and results are presented in section \ref{exp}. Finally, section \ref{conclusion} concludes the paper.

\section{Related Work}
\label{related_work}

%Ecologists are using computer vision and machine learning to develop novel techniques with various purposes.Ferreira et al. \cite{Ferreira2019DeepLM} trained a Convolutional Neural Network (CNN) with over three million images to recognize 48 different African animal species. They also developed an approach to efficiently gather labeled data and train networks to recognize small bird species. Additionally, 
Despite major advances in computer vision and machine learning, only a few studies have explored the potential of artificial intelligence for ecological applications similar to ours. In this context, Correia et al.\cite{Correia2020LeveragingAI} proposed a computer vision data processing system to detect open tree buds from time-lapse cameras for automating some tasks in phenological studies. They use a combination of a random forest, a CNN, and clustering to achieve detection. Kennedy et al. \cite{Kennedy2020AssessmentOL} presented a method using a fully connected feed-forward neural network for the assessment of lichen cover based on Landsat images, elevation data, and climatic parameters. However, their method does not detect lichens precisely. Instead, it directly regresses the percentage of lichen cover as a continuous value. Jozdani et al. \cite{Jozdani2021LeveragingDN} and Fraser et al. \cite{Fraser2022UAVAH} investigated the ability to train neural network models on high-resolution images taken with unmanned aerial drones. The method in \cite{Jozdani2021LeveragingDN} is designed to perform a binary segmentation of terricolous lichens, regardless of their species. Fraser et al. \cite{Fraser2022UAVAH} used a random forest model to globally quantify the cover of pale and fruticose lichens of the genus \textit{Cladonia} from UAV and satellite images. 

To identify changes in lichen populations, AI-based algorithms must be tested and refined in real-world contexts. Linear regression models showed that epiphytic lichen abundance was highly and positively correlated with the number of growth forms at all the geographical levels considered \cite{Aragn2019UsingGF}. However, previous works on lichen monitoring are mostly limited to assessing the overall lichen coverage globally from areal images. To the best of our knowledge, this is the first work to introduce an automated approach for monitoring individual lichen instances over extended periods, which provides a more fine-grained data analysis to support ecology research.

\section{Proposed Method}
\label{method}
In this work, we aim to automate the monitoring of epiphytic lichens by efficiently segmenting and distinguishing them in time-lapse images. Instance segmentation is required for the monitoring since it involves identifying and segmenting individual epiphytic lichens present on trees to estimate their change in size, which is directly related to lichen biomass. A bounding box is not precise enough. Therefore, we designed a method that relies on the use of an instance segmentation method to detect each separate object within the same category, and assign a unique label to their associated pixels. This allows us to precisely segment each individual epiphytic lichen in the image so that they can be further analyzed for their properties and distributions. For instance segmentation, we opted for Mask Scoring R-CNN \cite{Huang2019MaskSR}, which is a state-of-the-art object detection and instance segmentation model building on the popular Mask R-CNN model \cite{He2017MaskR}. It adds an extra branch to the Mask R-CNN network, which predicts a mask quality score for each detected object. This score is then used to adjust the mask probability before the final segmentation.

%\subsection{Lichen %segmentation}

%Our method is based on %Mask Scoring R-CNN %\cite{Huang2019MaskSR} %to perform instance %segmentation of %lichens individually. 
Compared to other state-of-the-art object detection and instance segmentation models, Mask Scoring R-CNN has demonstrated good performance on various benchmark datasets \cite{Zhang2016IsFR, Huang2019MaskSR, Harid2021AutomatedLE, Tu2020InstanceSB}. In addition, the model is highly customizable, allowing us to fine-tune its parameters and architecture for our specific application. It is an extension of the Mask R-CNN \cite{He2017MaskR} framework, by adding a Mask IoU head. The head component improves the quality of the predicted masks. It learns the quality of masks via regression, measured with a Mask IoU score, defined by
\begin{equation}
    MaskIoU=\frac{\textrm { Area of Intersection} }{\textrm { Area of Union }}
\end{equation}
and then penalizes the instance mask score if the classification score is high, while the actual mask
quality, given by $MaskIoU$, is low. More specifically, the inputs of this head are the predicted mask and a concatenated region of interest feature map. The Mask Scoring R-CNN loss is customized to optimize the segmentation quality and it is expressed as
\begin{equation}
    L=L_{cls}+L_{bbox}+L_{mask-scoring}+\lambda^*L_{mask-iou},
\end{equation}
where $L_{cls}$ represents the classification loss, $L_{bbox}$ represents the bounding box regression loss, $L_{mask-scoring}$ represents the mask IoU loss. $\lambda$ is a scalar weight for the mask scoring term and $L_{mask-iou}$ is the mask scoring term, which is defined as the average IoU between the predicted mask and the ground-truth mask for each object in the image.

Given a new Lichen time-lapse image, feature maps are constructed in the first stage by extracting image features of various scales using the backbone network. This is followed by the Region Proposal Network (RPN), which proposes candidate object regions, and the ROIAlign module, which extracts features for each region in the second stage. The resulting features are then fed into two parallel branches, where the first is for object detection and the second for instance segmentation, to predict the class, location, and binary mask of each lichen instance. Finally, a mask quality score is calculated for each predicted mask by considering the similarity to the ground-truth using a mask IoU branch, which is then combined with the original object detection score to produce the final score for the detected lichen instance.

\section{Experiments}
\label{exp}
\subsection{Dataset Construction}
The images in our dataset were collected for three different lichen species found in Canadian forests. Figure \ref{fig:lichens} shows an image of each lichen species. The images of the \textit{Erioderma pedicellatum} and \textit{Pectenia plumbea} are from the South coast of Newfoundland on the territory of Miawpukek First Nation. \textit{Erioderma pedicellatum} was growing on \textit{Abies balsamea} (L.) P. Mill. Tree. \textit{Pectenia plumbea} was growing on a \textit{Populus tremuloides} Michx. Tree.  The ones from \textit{Lobaria pulmonaria} are from an experiment on the west coast of Newfoundland and were growing on an Abies balsamea (L.) P. Mill. Tree.

The images were collected using a network of time-lapse cameras deployed by the Canadian Forest Service (CFS). The installed cameras capture an image every 2 hours. We removed blurry and dark images, as well as those mostly occluded due to snowfall. The composition of the resulting dataset used in our study is presented in Table \ref{tab:tab1}.

\begin{table}[t]
 \centering
 \small
  \caption{Description of  our data}
\begin{tabular}{|c|c|}
\hline \textbf{Lichen Type} & \textbf{Number of images} \\
\hline \textit{Pectenia plumbea (PP)} & 401 \\
\hline \textit{Erioderma pedicellatum (EP)} & 406 \\
\hline \textit{Lobaria pulmonaria (LP)} & 400 \\
\hline 
\hline 
Total & 1207 \\
\hline
\end{tabular}
\label{tab:tab1}%
\end{table}%

\begin{table*}[ht]
  \centering
  \caption{Data distribution of the \emph{cross-validation over lichen species} experiments. PP stands for \textit{Pectenia plumbea}, EP stands for \textit{Erioderma pedicellatum} and LP stands for \textit{Lobaria pulmonaria}. }
  \begin{tabular}{|c|c|c|c|}
    \hline
    & Training Data & Validation Data & Testing Data \\
    \hline
    Fold 1 & 686 images (341 images of \textit{PP} and 345 of \textit{EP}) & 121 images (60 images of \textit{PP} and 61 of \textit{EP}) & 400 images of \textit{LP} \\
    \hline
    Fold 2 & 681 images (341 images of \textit{PP} and 340 of \textit{LP}) & 120 images (60 images of \textit{PP} and 60 of \textit{LP}) & 406 images of \textit{EP} \\
    \hline
    Fold 3 & 685 images (345 images of \textit{EP} and 340 of \textit{LP}) & 121 images (61 images of \textit{EP} and 60 of \textit{LP}) & 401 images of \textit{PP} \\
    \hline
  \end{tabular}
  \label{tab:tab5}
\end{table*}

The regions of interest were carefully investigated, and the ground-truth of the lichens was manually annotated using the open-source VGG Image Annotator (VIA) tool \cite{Dutta2019TheVA}. The time-lapse images were then processed using data augmentation techniques, including random cropping, flipping, and rotation, before being fed into the local instance segmentation method for training. 

\subsection{Evaluation Metrics}
To evaluate the performance of our method, we used the mean average precision (mAP). This measure is calculated according to the following formula: 
\begin{equation}
    mAP=\frac{1}{9} \sum_{IoU \in \{0.5, \ldots, 0.95\}} AP_{IoU} ,
\end{equation}
with
\begin{equation}
AP_{IoU}=\frac{1}{9} \sum_{r \in \{0.5, \ldots, 0.95\}} \max_{\tilde{r} \geq r} p(\tilde{r}),
\end{equation}
where $AP_{IoU}$ is calculated as the mean of the precision $p$ for each recall value $r$ between 0.5 and 0.95, by adding steps of 0.05. $mAP50$ is the $mAP$ calculated for predicted masks that have an IoU with ground-truth masks of 50\% or more, whereas true positives in $mAP75$ must have an IoU with ground-truth masks of at least 75\%. Precision and recall are calculated as:
\begin{equation}
    Precision =\frac{TP}{TP+FP} ; \quad Recall =\frac{TP}{T P+FN} ;
\end{equation}
where $TP$ is the number of true positives, $FP$ is the number of false positives, and $FN$ is the number of false negatives. 
% Table generated by Excel2LaTeX from sheet 'Feuil1'

\subsection{Experimental approach and training}

\begin{figure}[t]
\centering
\includegraphics[width=0.52\textwidth]{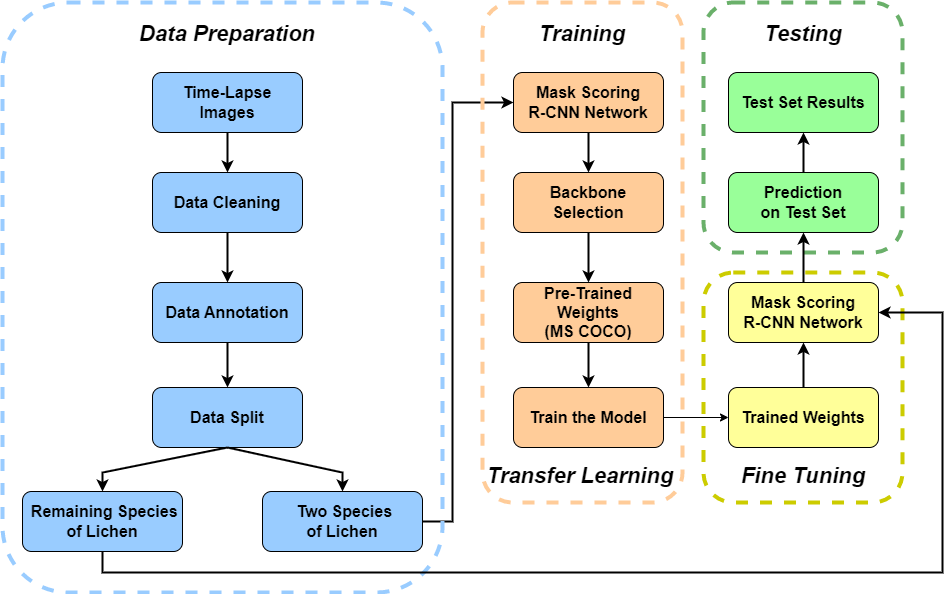}
\caption{Flowchart of our experimental approach}
\label{fig:flowchart}
\end{figure}

 We tested our method under three test scenarios. The flowchart of our experimental approach is presented in Figure \ref{fig:flowchart}. The goal is to build generic models capable of segmenting multiple types of lichens. We evaluate our method with scenarios with limited data because ecologists would then only be asked to manually segment a small portion of the data in order to automate the segmentation process for much larger datasets. The three scenarios are:
 
 \begin{enumerate}[i)]
     \item  \emph{Cross-validation over lichen species.} In this scenario, we used transfer learning to reduce the amount of training data required. We utilized pretrained weights from the Common Objects in Context (COCO) \cite{Lin2014MicrosoftCC} dataset to initialize all layers of our network, including the region proposal network (RPN), classifier, and mask head. This allowed us to leverage the features learned from the COCO dataset and adapt them to our specific task. We trained three models, each on two lichen species, to then test on the remaining species.  In this manner, we evaluate the model ability to recognize and segment a completely unknown lichen species that was not seen during training.

 \item \emph{Fine-Tuning on new lichen species.} In this scenario, we used the same models trained in the \emph{cross-validation over lichen species} scenario and added a fine-tuning step to the new species using a small amount of data selected at random.

 \item \emph{Selective Fine-Tuning on new lichen species.}  In this scenario, we used the same models trained in the \emph{cross-validation over lichen species} scenario and added a fine-tuning step to the new species using a particular subset of lichen species, that is one image per day of capture.
 \end{enumerate}

For the last two scenarios, we freeze the entire backbone during the fine-tuning because it has already been trained to generate a feature map for our specific problem, and resume training for the bounding box head and the mask head. These scenarios are designed to explore the potential of fine-tuning previously trained models on new unseen species with limited data

\begin{table}[b]
 \centering
  \caption{Results of the \emph{cross-validation over lichen species} experiments. We provide bounding box results (bbox) and segmentation results (segm): $ML_{PP-EP}$ corresponds to a model trained on PP and EP species. $ML_{PP-LP}$ corresponds to a model trained on PP and LP species. $ML_{EP-LP}$ corresponds to a model trained on EP and LP species.} 
\begin{tabular}{|c|c|c|c|c|c|c|c|}
\hline
  \multirow{2}{*}{} &\multicolumn{3}{|c|}{ bbox } & \multicolumn{3}{c|}{ segm } \\
  %\hline
  \cline{2-7}  
  & mAP & mAP50 & mAP75 & mAP & mAP50 & mAP75    \\
\hline  
$ML_{PP-EP}$  & 0.787 & 0.950 & 0.891 & 0.607 & 0.970 & 0.775  \\
\hline  
 $ML_{PP-LP}$ & 0.000 & 0.001 & 0.000 & 0.000 & 0.000 & 0.000 \\
\hline  
 $ML_{EP-LP}$ & 0.002 & 0.010 & 0.000 & 0.003 & 0.010 & 0.000 \\
\hline
\end{tabular}
\label{tab:tab6}%
\end{table}%
We evaluated different backbones for training the instance segmentation model and found that the best performance was achieved using a ResNet-50 backbone. We trained the models for 24 epochs using a learning rate of 0.05 and a batch size of 4, with a training schedule of 2x, which involves training the model for a certain number of epochs and then decreasing the learning rate by a factor of 10 and continuing training for another set of epochs.

\subsection{Results for cross-validation over lichen species}
The goal of this set of experiments is to evaluate the capacity of our method to segment a lichen species not included in the training data. To assess the effectiveness of this approach, a training process was conducted wherein three models were trained on data, each from two specific species. After the training process was completed, the performance of the models was evaluated by testing them directly on time-lapse images from the remaining third species. Table \ref{tab:tab5} presents the data used for each experiment performed during this process, including training, validation, and testing datasets.

Table \ref{tab:tab6} gives the results of the different experiments conducted, detailing the resulting mAP scores achieved by the models. This table shows that the performance of two models ($ML_{PP-LP}$ and $ML_{EP-LP}$) was low when tested on unrecognized types compared to $ML_{PP-EP}$. The experiments conducted provides evidence that the models were not capable of accurately segmenting certain species of epiphytic lichens that had not been trained on. The findings indicate that the $ML_{PP-LP}$, which was trained on PP and LP species, is not suitable for accurately segmenting EP lichen species. Similarly, $ML_{EP-LP}$ trained on EP and LP species is not effective for accurately segmenting PP lichen species.
Overall, these results indicate that some form of fine-tuning is required to improve the ability of these models to identify unknown lichen species, as simple transfer learning is not sufficient. 
\begin{table}[t]
\centering
\caption{Data distribution of the \emph{fine-tuning on new lichen species} experiments. PP stands for \textit{Pectenia plumbea}, EP stands for \textit{Erioderma pedicellatum} and LP stands for \textit{Lobaria pulmonaria}.}
\begin{tabular}{|c|c|c|c|}
\hline
           & Fine-tuning Data & Validation Data &  Testing Data \\
\hline
Fold 1 & 40 images of \textit{LP} & 10 images of \textit{LP} & 350 images of \textit{LP} \\
\hline
Fold 2 & 40 images of \textit{EP} & 10 images of \textit{EP} & 356 images of \textit{EP} \\
\hline
Fold 3 & 40 images of \textit{PP} & 10 images of \textit{PP} & 351 images of \textit{PP} \\
\hline
\end{tabular}  
\label{tab:tab70}%
\end{table}%

\begin{table}[b]
 \centering
  \caption{Cross-validation results of the \emph{fine-tuning on new lichen species} experiments. We provide bounding box results (bbox) and segmentation results (segm). We Fine-tuned the models obtained from the previous experiment.}
\begin{tabular}{|c|c|c|c|c|c|c|}
\hline
  \multirow{2}{*}{\begin{tabular}{@{}c@{}}Fine-tuned \\ model\end{tabular}} & \multicolumn{3}{c|}{bbox} & \multicolumn{3}{c|}{segm} \\

\cline{2-7}  
 & mAP & mAP50 & mAP75 & mAP & mAP50 & mAP75 \\
\hline 
\textbf{$ML_{PP-EP}$}  &  0.865 & 0.990 & 0.949 & 0.786 & 0.990 & 0.980 \\
\hline 
\textbf{$ML_{PP-LP}$}   & 0.679 & 0.980 & 0.814 & 0.585 & 0.977 & 0.662 \\
\hline 
\textbf{$ML_{EP-LP}$}   & 0.633 & 0.952 & 0.730 & 0.656 & 0.971 & 0.733 \\
\hline
\end{tabular}
\label{tab:tab71}%
\end{table}%

\begin{table}[ht]
 \centering
  \caption{Data distribution of the \emph{Selective Fine-tuning on new lichen species} experiments. PP stands for \textit{Pectenia plumbea}, EP stands for \textit{Erioderma pedicellatum} and LP stands for \textit{Lobaria pulmonaria}.}
\begin{tabular}{|c|c|c|c|}
\hline
           & Fine-tuning Data & Validation Data &  Testing Data \\
\hline
Fold 1 & 34 images of \textit{LP} & 12 images of \textit{LP} & 354 images of \textit{LP} \\
\hline
Fold 2 & 146 images of \textit{EP} & 47 images of \textit{EP} & 213 images of \textit{EP} \\
\hline
Fold 3 & 73 images of \textit{PP} & 24 images of \textit{PP} & 304 images of \textit{PP} \\
\hline
\end{tabular}  
\label{tab:tab7}%
\end{table}%

\begin{table}[b]
 \centering
  \caption{Cross-validation results of the \emph{Selective Fine-tuning on new lichen species} experiments. We provide bounding box results (bbox) and segmentation results (segm). We Fine-tuned the models obtained from the previous experiment.}
\begin{tabular}{|c|c|c|c|c|c|c|}
\hline
 \multirow{2}{*}{\begin{tabular}{@{}c@{}}Fine-tuned \\ model\end{tabular}} & \multicolumn{3}{c|}{bbox} & \multicolumn{3}{c|}{segm} \\

\cline{2-7} 
 &  mAP & mAP50 & mAP75 & mAP & mAP50 & mAP75 \\
 \hline 
\textbf{$ML_{PP-EP}$}  &  0.920 & 0.990 & 0.989 & 0.864 & 0.990 & 0.989 \\
\hline 
\textbf{$ML_{PP-LP}$}   & 0.858 & 0.988 & 0.958 & 0.768 & 0.982 & 0.952 \\
\hline 
\textbf{$ML_{EP-LP}$}   & 0.989 & 0.990 & 0.990 & 0.903 & 0.990 & 0.990 \\
\hline
\end{tabular}
\label{tab:tab8}%
\end{table}%

\subsection{Results of fine-tuning on new lichen species}
The goal of this set of experiments is to investigate the potential of using previously trained models ($ML_{PP-EP}$, $ML_{PP-LP}$, and $ML_{EP-LP}$) on new unknown species, once they have been fine-tuned on the target species using a limited random amount of data. In other words, by performing fine-tuning on the new species, we aimed to determine the feasibility of transferring the knowledge gained from training on multiple species to a new unseen species. This approach has several practical applications, such as enhancing the accuracy of species segmentation and minimizing the time and effort required for data annotation. This is highly advantageous in real-world scenarios, where ecologists can leverage the benefits of automated segmentation, by labeling only a small proportion of the available data. The data used for each experiment carried out during this process, including the fine-tuning, validation, and testing datasets are described in Table \ref{tab:tab70}.

\begin{table*}[t]
 \centering
  \caption{Comparative average results of the three experiments conducted in our study. We provide bounding box results (bbox) and segmentation results (segm). Best results are in bold}
\begin{tabular}{|c|c|c|c|c|c|c|}
\hline
 \multirow{2}{*}{Experiments} &  \multicolumn{3}{|c|}{ bbox } & \multicolumn{3}{c|}{ segm } \\
\cline{2-7} 
 &  mAP & mAP50 & mAP75 & mAP & mAP50 & mAP75 \\
 \hline 
\emph{Cross-validation over lichen species}  &  0.263 & 0.320 & 0.297 & 0.203 & 0.327 & 0.258 \\
\hline 
\emph{Fine-tuning on new lichen species}   & 0.726 & 0.974 & 0.831 & 0.676 & 0.979 & 0.792 \\
\hline 
\emph{Selective Fine-tuning on new lichen species}   & \textbf{0.922} & \textbf{0.989} & \textbf{0.979} & \textbf{0.845} & \textbf{0.987} & \textbf{0.977} \\
\hline
\end{tabular}
\label{tab:tab10}%
\end{table*}%

The results presented in Table \ref{tab:tab71} show that fine-tuning on new species can significantly improve the performance of lichen segmentation. These results indicate that even with minimal data, fine-tuning can have a substantial impact on the model ability to identify new species of lichens. The underlying assumption behind this experiment was that, despite appearance variations in epiphytic lichen species, they share certain characteristics that can be leveraged by pre-trained models. For example, they may have similar textures or colors that the model can recognize. Fine-tuning pre-trained models on a small amount of data from a new species helps to learn these new characteristics, which are specific to that species, leading to a significant segmentation quality improvement.

% Table generated by Excel2LaTeX from sheet 'Feuil1'

\subsection{Results of selective fine-tuning on new lichen species}
To further explore the potential of our models to segment new lichen species, we performed selective fine-tuning experiments. This involved taking the models that were obtained from the first set of experiments ($ML_{PP-EP}$, $ML_{PP-LP}$, and $ML_{EP-LP}$) and fine-tuning them selectively on a particular subset of epiphytic lichen. The fine-tuning subset includes one image from each day, that had not been included in the initial training data. By selecting only one image from each day, we are aiming to sample the diversity of lichen species over time. Additionally, selecting one image per day from every time-lapse camera ensures that the fine-tuned models are not biased toward any particular day or environmental conditions. During this experiment, we carefully monitored the effects of selective fine-tuning on a specific subset. The data used for each experiment carried out during this process, including the fine-tuning, validation, and testing datasets is detailed in Table \ref{tab:tab7}.

The results of \emph{selective fine-tuning on new lichen species} experiments, shown in table \ref{tab:tab8}, demonstrate even more the effect of fine-tuning on a specific type of lichen. Through a systematic sampling process consisting of taking a single image from each day, we were able to observe the impact of selective fine-tuning on the model mAP scores and draw conclusions regarding its potential for segmenting a broader range of lichen species. The systematic sampling approach used in this experiment proved to be more effective than random sampling with the previous experiment (\emph{fine-tuning on new lichen species}). We also see that selective fine-tuning improves the model ability to detect lichens with higher Intersection over Union (IoU) scores, as demonstrated by the significant improvement in mAP75 compared to mAP50. Therefore, selective fine-tuning is particularly effective for more challenging cases where a higher IoU score is required.

\subsection{Discussion}

Table \ref{tab:tab10} presents a comparison of the average results obtained from the experiments previously conducted. We can conclude from this table that selective fine-tuning of the model using a single image from each day is demonstrated to be an effective strategy. This suggests that environmental conditions play an important role in the model performance. In fact, diverse weather and lighting conditions resulting from choosing a single image per day led to improved mAP scores. Therefore, it is important to carefully consider the environmental factors when designing and training models for lichen species segmentation.
\section{Conclusion}
\label{conclusion}
In this paper, we designed a deep learning framework for segmenting epiphytic lichens, which represents the first computer vision method for automating the monitoring of epiphytic lichens using time-lapse cameras. The ability of our model to recognize different species of epiphytic lichens and track their progress over subsequent seasons from limited annotated data makes it a comprehensive and efficient approach for long-term, large-scale ecological monitoring. Our approach has a great potential to assist ecologists in identifying and tracking changes in lichen populations, and thus understanding the impact of climate change on forests. 
\begin{comment}
As a future work, we aim to expand our approach to also detect and track the reproductive structures of lichens, which also represent an important ecological indicator. 
These reproductive structures are often visible as small, brightly colored dots on the lichen thallus. Detecting and tracking the evolution of lichen reproductive structures can provide valuable bioindicators and information about changes in the environment, such as air pollution, climate change, and habitat disturbance. 
\end{comment}

\section*{Acknowledgment}

The authors would like to thank the Miawpukek Forest guardians, Andy Joe, Raymond Jeddore, David Jeddore, and Greg Benoit who assisted in many parts of the project in the field and shared their knowledge of the land. We would also acknowledge support from the NRCAN’s ADM innovation fund, the CFS Sustainable Forest Management program, and the Natural Sciences and Engineering Research Council of Canada (NSERC), [NSERC funding references: RGPIN-2020-04633 and RGPIN-2020-04937].

\bibliographystyle{IEEEtran}
\bibliography{bibliography}

\end{document}